\documentclass[conference]{IEEEtran}
\usepackage{times}

\usepackage[numbers]{natbib}
\usepackage{multicol}
\usepackage[bookmarks=true]{hyperref}

\usepackage{amsfonts}       
\usepackage{nicefrac}       
\usepackage{microtype}      
\usepackage{graphicx}
\usepackage{subcaption}
\usepackage[dvipsnames]{xcolor}
\usepackage{wrapfig}
\usepackage{amssymb,amsmath}
\usepackage{etoolbox}
\usepackage[ruled]{algorithm2e}
\usepackage{algpseudocode}
\usepackage{multirow}
\usepackage{hhline}
\let\labelindent\relax
\usepackage{enumitem}

\usepackage{comment}
\usepackage{fancybox}
\usepackage{pbox}
\usepackage{tcolorbox}
\usepackage{dsfont}
\usepackage{capt-of}
\usepackage{bm}
\usepackage{array}
\usepackage{multirow}
\usepackage{tikz}
\usepackage{scalerel}
\usepackage{siunitx}
\usepackage[T1]{fontenc}
\usepackage{booktabs}
\newcommand{\obs}{x}
\newcommand{\act}{a}

\newcommand{\qh}{h^{\scaleto{Q}{5pt}}}
\newcommand{\pih}{h^{\pi}}

\newcommand{\paramr}{\psi}
\newcommand{\paramc}{\theta}
\newcommand{\parama}{\phi}

\newcommand{\lift}{{\small \texttt{lift\_green}}}
\newcommand{\liftcloth}{{\small \texttt{lift\_cloth}}}
\newcommand{\stack}{{\small \texttt{stack\_green\_on\_red}}}
\newcommand{\randwatch}{{\small \texttt{random\_watcher}}}
\newcommand{\rgb}{\emph{rgb}}
\newcommand{\deform}{\emph{deformable}}
\newcommand{\Rgb}{\emph{RGB}}
\newcommand{\Deform}{\emph{Deformable}}

\newcommand{\citetemp}[1]{\textcolor{red}{[X]}}

\hypersetup{
    colorlinks=true,
    linkcolor=black,
    urlcolor=blue,
}

\newcommand{\neverendingstorage}{NES}
\newcommand{\Neverendingstorage}{NES}

\newcommand{\be}{\begin{equation}}
\newcommand{\ee}{\end{equation}}
\newcommand{\bea}{\begin{eqnarray}}
\newcommand{\eea}{\end{eqnarray}}
\newcommand{\beaa}{\begin{eqnarray*}}
\newcommand{\eeaa}{\end{eqnarray*}}

\DeclareMathAlphabet{\mathpzc}{OT1}{pzc}{m}{n}

\title{\LARGE \bf Scaling data-driven robotics with reward sketching and \protect\\ batch reinforcement learning}

\author{
\authorblockN{Serkan Cabi,
Sergio G{\'o}mez Colmenarejo,
Alexander Novikov,
Ksenia Konyushkova, \\
Scott Reed,
Rae Jeong,
Konrad {\.Z}o{\l}na,
Yusuf Aytar,
David Budden,
Mel Vecerik, \\
Oleg Sushkov,
David Barker,
Jonathan Scholz,
Misha Denil,
Nando de Freitas,
Ziyu Wang}
\authorblockA{Deepmind}
\thanks{This work is done at Deepmind.
Videos, dataset and agent code are available through this website: \href{https://sites.google.com/view/data-driven-robotics/}{https://sites.google.com/view/data-driven-robotics/}}
}

\begin{document}

\maketitle
\thispagestyle{empty}
\pagestyle{empty}


\begin{abstract}

By harnessing a growing dataset of robot experience, we learn control policies for a diverse and increasing set of related manipulation tasks. To make this possible, we introduce reward sketching: an effective way of eliciting human preferences to learn the reward function for a new task.
This reward function is then used to retrospectively annotate all historical data, collected for different tasks, with predicted rewards for the new task.
The resulting massive annotated dataset can then be used to learn manipulation policies with batch reinforcement learning (RL) from visual input in a completely off-line way, {\emph i.e.}, without interactions with the real robot. This approach makes it possible to scale up RL in robotics, as we no longer need to run the robot for each step of learning.
We show that the trained batch RL agents, when deployed in real robots, can perform a variety of challenging tasks involving multiple interactions among rigid or deformable objects. Moreover, they display a significant degree of robustness and generalization. In some cases, they even outperform human teleoperators.

\end{abstract}

\section{Introduction}
\label{sec:intro}

Deep learning has successfully advanced many areas of artificial intelligence, including vision~\cite{krizhevsky2012imagenet,he2016deep}, speech recognition~\cite{graves2014towards,maas2015lexicon,amodei2016deep}, natural language processing~\cite{devlin2019bert}, and reinforcement learning (RL)~\cite{mnih2015human,silver2016mastering}.
The success of deep learning in each of these fields was made possible by the availability of huge amounts of labeled training data.
Researchers in vision and language can easily train and evaluate deep neural networks on standard datasets with crowdsourced annotations such as ImageNet~\cite{ILSVRC15}, COCO~\cite{lin2014microsoft} and CLEVR~\cite{johnson2017clevr}.
In simulated environments like video games, where experience and rewards are easy to obtain, deep RL is tremendously successful in outperforming top skilled humans by ingesting huge amounts of data~\cite{silver2016mastering, vinyals2019alphastar, berner2019dota}.
The OpenAI Five DOTA bot~\cite{berner2019dota} processes 180 years of simulated experience every day to play at a professional level.
Even playing simple Atari games typically requires 40 days of game play \cite{mnih2015human}.
In contrast, in robotics we lack abundant data since data collection implies execution on a real robot, which cannot be accelerated beyond real time.
Furthermore, task rewards do not naturally exist in the real-world robotics as it is the case in simulated environments.
The lack of large datasets with reward signals has limited the effectiveness of deep RL in robotics.

\begin{figure}[t]
\centering
  \includegraphics[width=0.96\columnwidth]{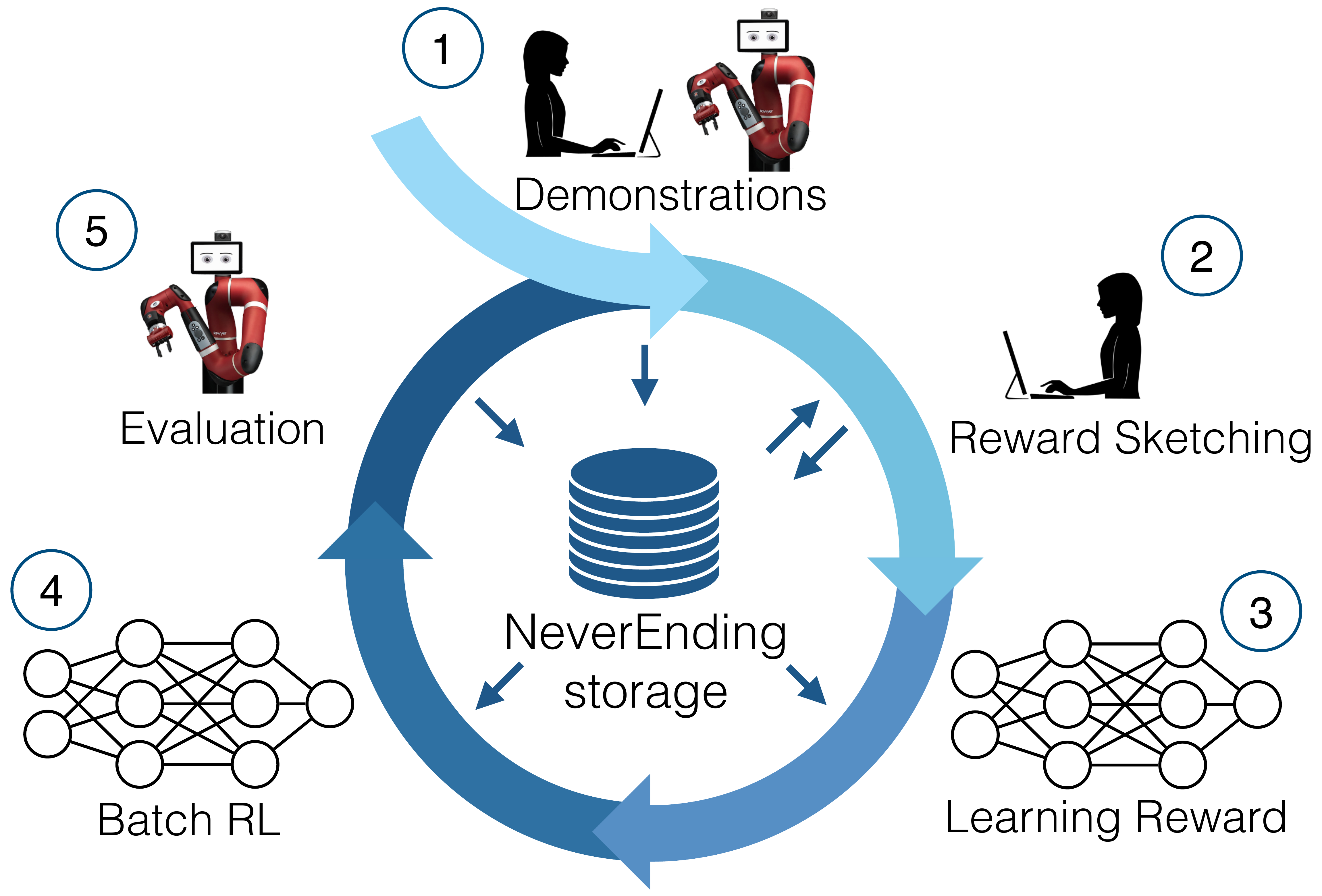}
\caption{Our cyclical approach for \emph{never-ending} collection of data and continual learning of new tasks consists of five stages: (1)~generation of observation-action pairs by either teleoperation, scripted policies or trained agents, (2)~a novel interactive approach for eliciting preferences for a specific new task, (3)~learning the reward function for the new task and applying this function to automatically label all the historical data, (4)~applying batch RL to learn policies purely from the massive growing dataset, without online interaction, and (5)~evaluation of the learned policies.
}
\vspace{-5mm}
\label{fig:pull-figure}
\end{figure}

This paper presents a data-driven approach to apply deep RL effectively to learn to perform manipulation tasks on real robots from vision.
Our solution is illustrated in \autoref{fig:pull-figure}.
At its heart are three important ideas: (i) efficient elicitation of user  preferences to learn reward functions, (ii) automatic annotation of all historical data  with any of the learned reward functions, and (iii) harnessing the large annotated datasets to learn policies purely from stored data via batch RL.

\begin{figure}[t]
	\centering
	\includegraphics[width=\columnwidth]{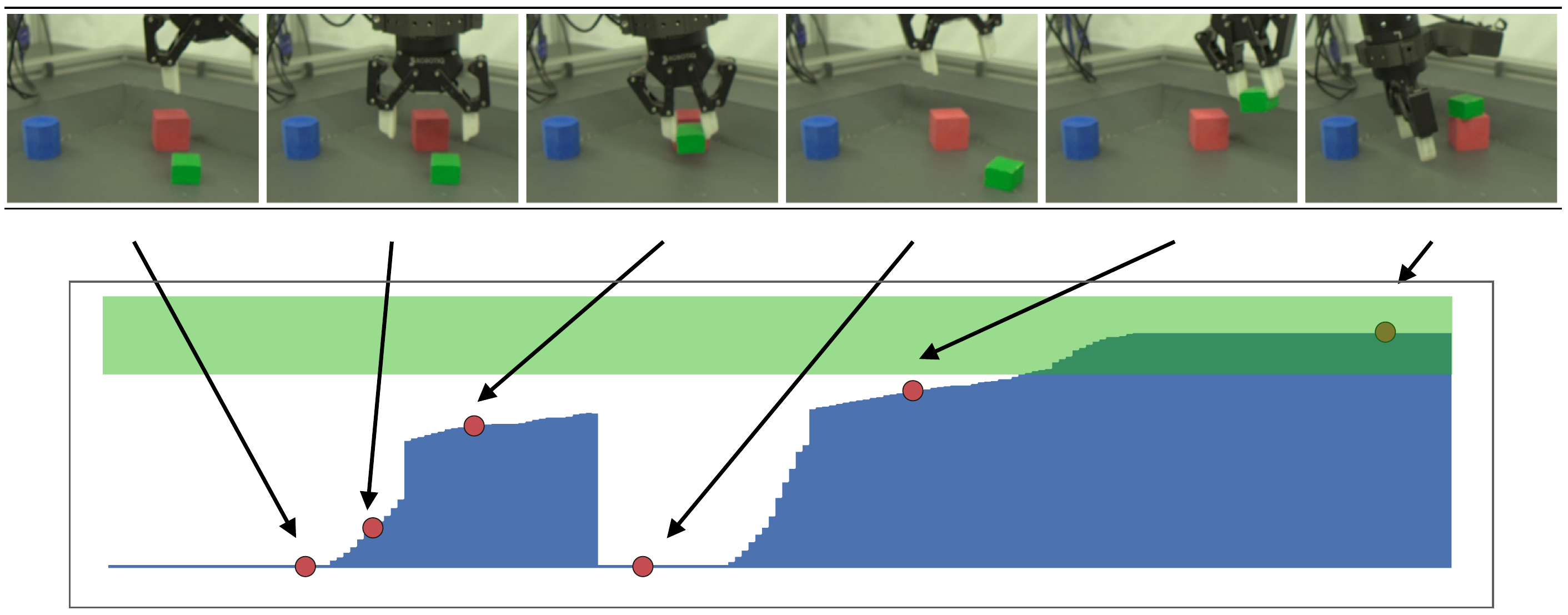}
	\caption{Reward sketching procedure. Sketch of a reward function for \stack{} task.
	A video sequence (top) with a {\em reward sketch} (bottom), shown in blue.
	Reward is the perceived progress towards achieving the target task.
	The annotators are instructed to indicate successful timesteps with reward high enough to reach green area.}
	\label{fig:interface}
	\vspace{-5mm}
\end{figure}

Existing RL approaches for real-world robotics mainly focus on tasks where hand-crafted reward mechanisms can be developed.
Simple behaviours such as learning to grasp objects~\cite{kalashnikov2018qtopt} or learning to fly~\cite{gandhi2017learning} by avoiding crashing can be acquired by reward engineering.
However, as the task complexity increases, this approach does not scale well.
We propose a novel way to specify rewards that allows to generate reward labels for a large number of diverse tasks.
Our approach relies on human judgments about progress towards the goal to train task-specific reward functions.
Annotations are elicited from humans in the form of per-timestep reward annotations using a process we call \emph{reward sketching}, see \autoref{fig:interface}.
The sketching procedure is intuitive for humans, and allows them to label many timesteps rapidly and accurately.
We use the human annotations to train a ranking reward model, which is then used to annotate all other episodes.

\begin{figure}[t]
	\centering
	\includegraphics[width=\columnwidth]{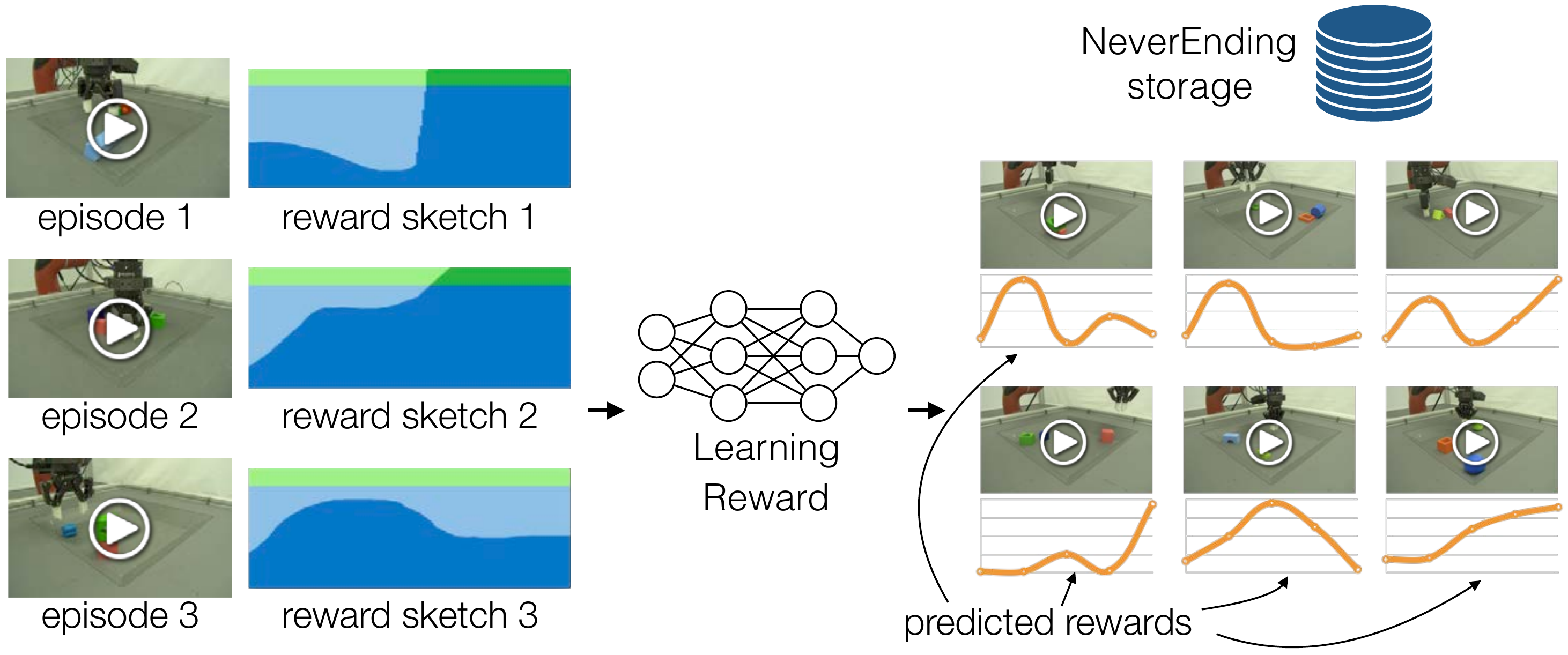}
	\caption{Retrospective reward assignment. The reward function is learned from a limited set of episodes with reward sketches. The learned reward function is applied to a massive dataset of episodes from NeverEnding Storage. All historical episodes are now labelled with a newly learned reward function.}
	\vspace{-5mm}
	\label{fig:retrospective}
\end{figure}

To generate enough data to train data-demanding deep neural network agents,
we record experience continuously and persistently, regardless of the purpose or quality of the behavior.
We collected over \num{400} hours of multiple-camera videos (\autoref{fig:rig}), proprioception, and actions from behavior generated by human teleoperators, as well as random, scripted and trained  policies.
By using deep reward networks obtained as a result of reward sketching, it becomes possible to retrospectively assign rewards to any past or future experience for any given task.
Thus, the learned reward function allows us to repurpose a \emph{large} amount of past experience using a \emph{fixed} amount of annotation effort per task, see \autoref{fig:retrospective}.
This large dataset with task-specific rewards can now be used to harness the power of deep batch RL.

For any given new task, our data is necessarily off-policy, and is
typically off-task (\emph{i.e.}, collected for other tasks).
In this case, batch RL~\cite{lange2012batch} is a good method to learn visuomotor policies.
Batch RL effectively enables us to learn new controllers without execution on the robot.
Running RL off-line gives researchers several advantages.
Firstly, there is no need to worry about wear and tear, limits of real-time processing, and many of the other challenges associated with operating real robots.
Moreover, researchers are empowered to train policies using their batch RL algorithm of choice, similar to how vision researchers are empowered to try new methods on ImageNet.
To this end, we release datasets~\cite{outdataset} with this paper.

The integration of all the elements into a scalable system to tightly close the loop of human input, reward learning and policy learning poses substantial engineering challenges.
Nevertheless, this work is essential to advance the data-driven robotics.
For example, we store all robot experience including demonstrations, behaviors generated by trained policies or scripted random policies.
To be useful in learning, this data needs to be appropriately annotated and queried.
This is achieved thanks to a design of our storage system dubbed NeverEnding Storage (\Neverendingstorage{}).

\begin{figure}[t]
\centering
  \includegraphics[width=1\columnwidth]{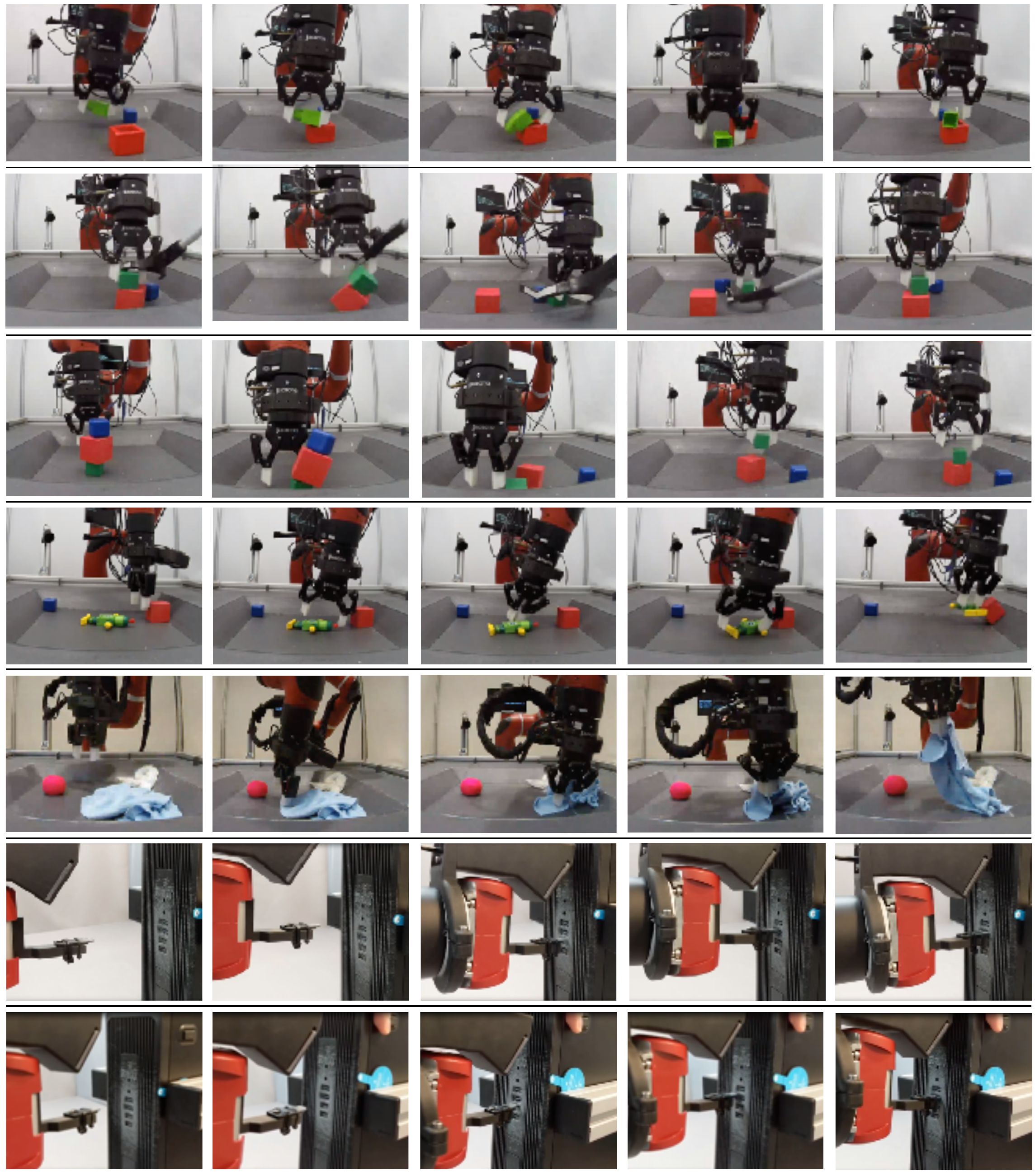}
\caption{Each row is an example episode of a successful task illustrating: (1) the ability to recover from a mistake in \stack{} task, (2) robustness to adversarial perturbations in the same task, (3) generalization to unseen initial conditions in the same task, 4) generalizing to previously unseen objects in a \lift{} task, (5) the ability to lift deformable objects, (6) inserting a USB key, (7) inserting a USB key despite moving the target computer.}
\vspace{-5mm}
\label{fig:film-strip}
\end{figure}

This multi-component system (\autoref{fig:pull-figure}) allows us to solve a variety of challenging tasks (\autoref{fig:film-strip}) that require skillful manipulation, involve multi-object interaction, and consist of many time steps.
An example of such task is stacking arbitrarily shaped objects.
In this task, small perturbations at the beginning can easily cause failure later:
The robot not only has to achieve a successful grasp, but it must also grasp the first object in a way that allows for safe placement on top of the second object.
Moreover, the second object may have a small surface area which varies how demanding the task is.
Learning policies directly from pixels makes the task more challenging, but eliminates the need for feature engineering and allows for additional generalization capacity.
While some of our tasks can be solved effectively with scripted policies, learning policies that generalize to arbitrary shapes, sizes, textures and materials remains a formidable challenge, and hence the focus of this paper is on making progress towards meeting this challenge.

As shown in \autoref{fig:film-strip}, the policies learned with our approach solve a variety of tasks including lifting and stacking of rigid/deformable objects, as well as USB insertion.
Importantly, thanks to learning from pixels, the behaviour generalizes to new object shapes and to new initial conditions, recovers from mistakes and is robust to some real-time adversarial interference.
\autoref{fig:agent_vs_human} shows that the learned policies can also solve tasks more effectively than human teleoperators.
\emph{To better view our results and general approach, we highly recommend watching the accompanying video on the \href{https://sites.google.com/corp/view/data-driven-robotics/}{\underline{project website}}.}

The remainder of this paper is organized as follows. \autoref{sec:method} introduces the methods, focusing on reward sketching, reward learning and batch RL, but also provides the bigger context highlighting the engineering contributions.
\autoref{sec:experiments} is devoted to describing our experimental setup, network architectures, benchmark results, and an interactive insertion task of industrial relevance.
\autoref{sec:relwork} explores some of the related work.

\section{Methods}
\label{sec:method}

\begin{figure}[t]
	\centering
    \includegraphics[width=\columnwidth]{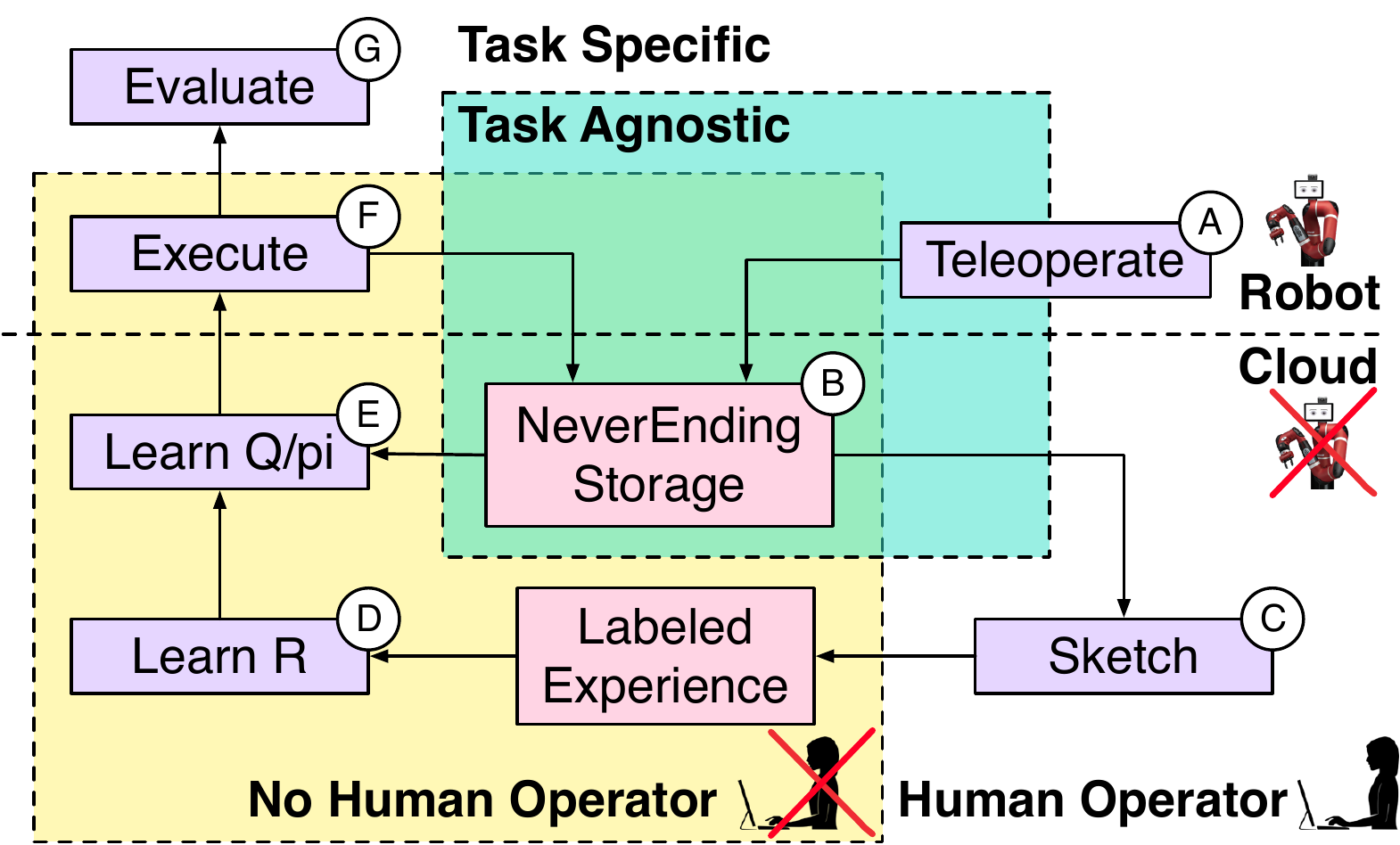}
    \caption{Structure of the data-driven workflow. Each step is described in \autoref{sec:method} and the figure highlights which steps are performed on the robot or not, involving human operator or not and if they are task-specific or task-agnostic.}
\vspace{-5mm}
	\label{fig:workflow}
\end{figure}

The general workflow is illustrated in \autoref{fig:pull-figure} and a more detailed procedure is presented in \autoref{fig:workflow}.
\neverendingstorage{} accumulates a large dataset of task-agnostic experience.
A task-specific reward model allows us to retrospectively annotate data in \neverendingstorage{} with reward signals for a new task.
With rewards, we can then train batch RL agents with all the data in \neverendingstorage{}.

The procedure for training an agent to complete a new task has the following steps which are described in turn in the remainder of the section:
\begin{enumerate}[label=\Alph*.]
    \item A human teleoperates the robot to provide first-person demonstrations of the target task.
    \item All robot experience, including demonstrations, is accumulated into \neverendingstorage{}.
    \item Humans annotate a subset of episodes from \neverendingstorage{} (including task-specific demos) with reward sketches.
    \item A reward model for the target task is trained using the fixed amount of labelled experience.
    \item An agent for the target task is trained using all experience in \neverendingstorage{}, using the predicted reward values.
    \item The resulting policy is deployed on a real robot, while recording more data into \neverendingstorage{}. can further be annotated.
    \item Occasionally we select an agent for careful evaluation, to track overall progress on the task.
\end{enumerate}

\subsection{Teleoperation}
\label{sec:method-detail:teleop}

To specify a new target task, a human operator first remotely controls the robot to provide several successful (and occasionally unsuccessful\footnote{We notice that in our dataset around $15$\% of human demonstrations fail to accomplish the task at the end of the episode.}) examples of completing the task.
By employing the demonstration trajectories, we facilitate both reward learning and reinforcement learning tasks.
Demonstrations help to bootstrap the reward learning by providing examples of successful behavior with high rewards, which are also easy to interpret and judge for humans.
In RL, we circumvent the problem of exploration: Instead of requiring that the agent explores the state space autonomously, we use expert knowledge about the intended outcome of the task to guide the agent.
In addition to full episodes of demonstrations, when an agent controls the robot, \emph{interactive interventions} can be also performed:
A human operator can take over from, or return control to, an agent at any time.
This data is useful for fixing particular corner cases that the agents might encounter.

The robot is controlled with a $6$-DoF mouse with an additional gripper button (see the video) or hand-held virtual reality controller.
A demonstrated sequence contains pairs of observations and corresponding actions for each time step $t$: $((\obs_0, \act_0), \ldots, (\obs_t, \act_t), \ldots, (\obs_T, \act_T))$.
Observations $\obs_t$ contain all available sensor data including raw pixels from multiple cameras as well as proprioceptive inputs (\autoref{fig:rig}).

\subsection{NeverEnding Storage}
\label{sec:details-nes}

\neverendingstorage{} captures all of the robot experience generated across all tasks in a central repository.
This allows us to make use of historical data each time when learning a new target task, instead of generating a new dataset from scratch.
\neverendingstorage{} includes teleoperated trajectories for various tasks, human play data, and experience from the execution of either scripted or learned policies.
For every trajectory we store recordings from several cameras and sensors in the robot cage (\autoref{fig:rig}).
The main innovation in \Neverendingstorage{} is the introduction of a rich metadata system into the RL training pipeline.
It is implemented as a relational database that can be accessed using SQL-type queries.
We attach environment and policy metadata to every trajectory (e.g., date and time of operation), as well as arbitrary human-readable labels and reward sketches.
This information allows us to dynamically retrieve and slice the data relevant for a particular stage of our training pipeline.

\subsection{Reward Sketching}
\label{sec:method-details:sketch}

The second step in task specification is \emph{reward sketching}.
We ask human experts to provide per-timestep annotations of reward using a custom user interface.
As illustrated in \autoref{fig:interface}, the user draws a curve indicating the progress towards accomplishing the target task as a function of time, while the interface shows the frame corresponding to the current cursor position.
This intuitive interface allows a single annotator to produce hundreds of frames of reward annotations per minute.

To sketch an episode, a user interactively selects a frame $\obs_t$ and provides an associated reward value $s(\obs_t) \in [0, 1]$.
The sketching interface allows the annotator to draw reward curves while ``scrubbing'' through a video episode, rather than annotating frame by frame.
This efficient procedure provides a rich source of information about the reward across the entire episode.
The sketches for an episode $\{s(\obs_t)\}|_{t=1}^{T}$ are stored in \neverendingstorage{} as described in~\autoref{sec:details-nes}.

The reward sketches allow comparison of perceived value of any two frames.
In addition, the green region in \autoref{fig:interface} is reserved for frames where the goal is achieved.
For each task the episodes to be annotated are drawn from \neverendingstorage{}.
They include both the demonstrations of the target task, as well as experience generated for prior tasks.
Annotating data from prior tasks ensures better coverage of the state space.

Sketching is particularly suited for tasks where humans are able to compare two timesteps reliably. Typical object manipulation tasks fall in this category, but not all robot tasks are like this. For instance, it would be hard to sketch tasks where variable speed is important, or with cycles as in walking.
While we are aware of these limitations, the proposed approach does however cover many manipulation tasks of interest as shown here. We believe future work should advance interfaces to address a wider variety of tasks.

\subsection{Reward Learning}
\label{sec:method:reward}

The reward annotations produced by sketching are used to train a reward model.
This model is then used to predict reward values for all experience in \neverendingstorage{} (\autoref{fig:retrospective}).
As a result, we can leverage all historical data in training a policy for a new task, without manual human annotation of the entire repository.

Episodes annotated with reward sketches are used to train a reward function in the form of neural network with parameters $\paramr$ in a supervised manner.
We find that although there is high agreement between annotators on the relative quality of timesteps within an episode, annotators are often not consistent in the overall \emph{scale} of the sketched rewards.
We therefore adopt an intra-episode ranking approach to learn reward functions, rather than trying to regress the sketched values directly.

Specifically, given two frames $\obs_t$ and $\obs_q$ in the same episode, we train the reward model to satisfy two conditions.
First, if frame $\obs_t$ is (un)successful according to the sketch $s(\obs_t)$, it should be (un)successful according the estimated reward function $r_{\paramr}(\obs_t)$.
The successful and unsuccessful frames in reward sketches are defined by exceeding or not a threshold $\tau_s$, the (un)successful frames in the predicted reward exceed (or not) a threshold $\tau_{r1}$ ($\tau_{r2}$).
Second, if $s(\obs_t)$ is higher than $s(\obs_q)$ by a threshold $\mu_s$, then
$r_{\paramr}(\obs_t)$ should be higher than $r_{\paramr}(\obs_q)$ by another threshold $\mu_r$.
These conditions are captured by the following two hinge losses:
\begin{align*}
    \mathcal{L}_{rank}(\paramr) =&
    \max \left\{0, r_{\paramr}(\obs_t)-r_{\paramr}(\obs_q) + \mu_r \right\}  \mathds{1}_{s(\obs_q)-s(\obs_t) > \mu_s}
    \nonumber \\
    \mathcal{L}_{success}(\paramr) =&
        \max \left\{0, \tau_{r1} - r_{\paramr}(\obs) \right\}  \mathds{1}_{s(\obs) > \tau_s } \; + \; \nonumber \\
        &\max \left\{0, r_{\paramr}(\obs) - \tau_{r2} \right\} \mathds{1}_{s(\obs) < \tau_s }
    \label{eq:success}
\end{align*}
The total loss is obtained by adding these terms: $\mathcal{L}_{rank}+ \lambda \mathcal{L}_{success}$.
In our experiments, we set $\mu_s = 0.2$, $\mu_r = 0.1$, $\tau_s=0.85$, $\tau_{r1}=0.9$, $\tau_{r2}=0.7$, and $\lambda=10$.

\subsection{Batch RL}
\label{sec:method-details:off-policy}

We train policies using batch RL \cite{lange2012batch}.
In batch RL, the new policy is learned using a single batch of data generated by different previous policies, and without further execution on the robot.
Our agent is trained using only distributional RL~\cite{barthmaron2018distributed}, without any feature pretraining, behaviour cloning (BC) initialization, any special batch correction terms, or auxiliary losses.
We do, however, find it important to use the historical data from other tasks.

Our choice of distributional RL is partly motivated by the success of this method for batch RL in Atari \cite{agarwal2019striving}. We compare the distributional and non-distributional RL alternatives in our experiments. We note that other batch RL methods (see \autoref{sec:relwork}) might also lead to good results. Because of this, we release our datasets~\cite{outdataset} and canonical agents~\cite{hoffman2020acme} to encourage further investigation and advances to batch RL algorithms in robotics.

We use an algorithm similar to D4PG~\cite{barthmaron2018distributed, hoffman2020acme} as our training algorithm.
It maintains a value network $Q(\obs_{t}, \qh_{t}, \act \,|\, \paramc)$ and a policy network $\pi(\obs, \pih_{t} \,|\, \parama)$.
Given the effectiveness of recurrent value functions \cite{kapturowski2018recurrent}, both $Q$ and $\pi$ are recurrent with $\qh_{t}$ and $\pih_{t}$ representing the corresponding recurrent hidden states.
The target networks have the same structure as the value and policy networks, but are parameterized by different parameters $\paramc'$ and $\parama'$, which are periodically updated to the current parameters of the original networks.

Given the $Q$ function, we update the policy using DPG~\cite{silver2014deterministic}.
As in D4PG, we adopt a distributional value function \cite{bellemare2017distributional} and minimize the associated loss to learn the critic.
During learning, we sample a batch of sequences of observations and actions
$\{\obs_t^i, \act_t^i, \cdots, \obs_{t+n}^i\}_{i}$ and
use a zero start state to initialize all recurrent states at the start of sampled sequences.
We then update $\parama$ and $\paramc$ using BPTT \cite{werbos1990backpropagation}.

Since \neverendingstorage{} contains data from many different tasks, a randomly sampled batch from \neverendingstorage{} may contain data mostly irrelevant to the task at hand.
To increase the representation of data from the current task, we construct fixed ratio batches, with \num{75}\% of the batch drawn from the entirety of \neverendingstorage{} and \num{25}\% from the data specific to the target task.
This is similar to the solution proposed in previous work~\cite{pohlen2018observe}, where fixed ratio batches are formed with agent and demonstration data.

\subsection{Execution}
\label{sec:method-details:execution}

Once an agent is trained, we can run it on the real robot.
By running the agent, we collect more experience, which can be used for reward sketching or RL in future iterations. Running the agent also allows us to observe its performance and make judgments about the steps needed to improve it.

In early workflow iterations, before the reward functions are trained with sufficient coverage of state space, the policies often exploit ``delusions'' where high rewards are assigned to undesired behaviors.
To fix a reward delusion, a human annotator sketches some of the episodes where the delusion is observed.
New annotations are used to improve the reward model, which is used in training a new policy.
For each target task, this cycle is typically repeated \num{2}--\num{3} times until the predictions of a reward function are satisfactory.

\section{Experiments}
\label{sec:experiments}

\subsection{Experimental Setup}
\label{sec:exp:setup}

\begin{figure}[t]
	\centering
	\includegraphics[width=0.96\columnwidth]{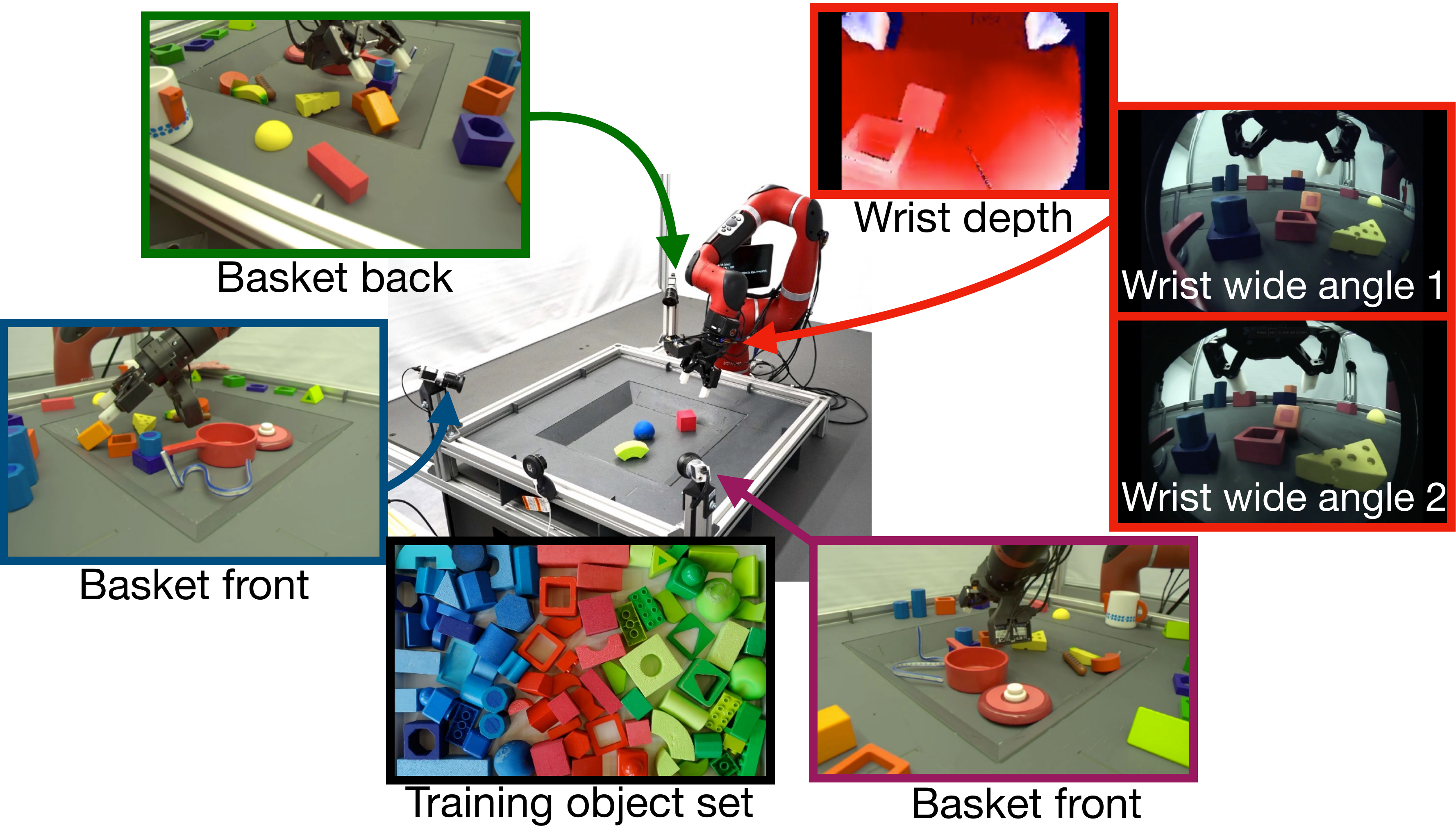}
	\caption{The robot senses and records all data acquired with its $3$ cage cameras, $3$ wrist cameras (wide angle and depth) and proprioception. It also records its actions continuously. The robot is trained with a wide variety of object shapes, textures and sizes to achieve generalization at deployment time.}
	\vspace{-5mm}
	\label{fig:rig}
\end{figure}

\emph{Robotic setup: }
Our setup consists of a Sawyer robot with a Robotiq \num{2}F-\num{85} gripper and a wrist force-torque sensor facing a \num{35x35} cm basket.
The action space has six continuous degrees of freedom, corresponding to Cartesian translational and rotational velocity targets of the gripper pinch point and one binary control of gripper fingers.
The agent control loop is executed at \num{10}Hz.
For safety, the pinch point movement is restricted to be in a \num{35x35x15} cm workspace with maximum rotations of \num{30}$^{\circ}$, \num{90}$^{\circ}$, and \num{180}$^{\circ}$ around each axis.

Observations are provided by three cameras around the cage, as well as two wide angle cameras and one depth camera mounted at the wrist, and proprioceptive sensors in the arm (\autoref{fig:rig}).
\neverendingstorage{} captures all of the observations, and we indicate what subset is used for each learned component.

\begin{figure*}[t]
	\centering
	\includegraphics[width=\textwidth]{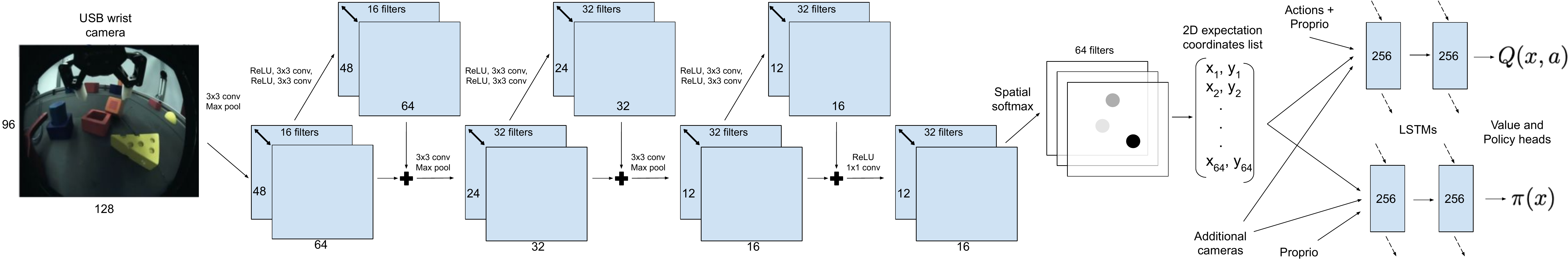}
	\caption{Agent network architecture. Only the wrist camera encoder is shown here, but in practice we encode each camera independently and concatenate the results.
	\label{fig:agent_network}}
	\vspace{-5mm}
\end{figure*}

\emph{Tasks and datasets: }
We focus on $2$ subsets of \neverendingstorage{}, with data recorded during manipulation of
$3$ variable-shape rigid objects coloured red, green and blue (\rgb{} dataset, \autoref{fig:rig}), and
$3$ deformable objects: a soft ball, a rope and a cloth (\deform{} dataset, \autoref{fig:film-strip}, row $5$).
The \rgb{} dataset is used to learn policies for two tasks: \lift{} and \stack{}, and the \deform{} dataset is used for the \liftcloth{} task.
Statistics for both datasets are presented in~\autoref{tab:dataset} which describes how much data is teleoperated, how much comes from the target tasks and how much is obtained by random scripted policies.
Each episode lasts for $200$ steps ($20$ seconds) unless it is terminated earlier for safety reasons.

\begin{table}[t]
\begin{subtable}{\columnwidth}
    \centering
        \begin{tabular}{>{\raggedright}p{3.4cm} | r | r | r}
        \toprule
        Type & No. Episodes & No. steps & Hours\\
        \midrule
        Teleoperation & 6.2 K & 1.1 M & 31.9 \\
        \lift{} & 8.5 K & 1.5 M & 41.3 \\
        \stack{} & 10.3 K & 2.0 M & 56.1 \\
        \randwatch{} & 13.1 K & 2.6 M & 70.9 \\
        Total & 37.9 K & 7.0 M & 193.3 \\
        \bottomrule
        \end{tabular}
    \caption{\Rgb{} dataset. \label{tab:rgb_dataset}}
\end{subtable}
\bigskip

\begin{subtable}{\columnwidth}
    \centering
        \begin{tabular}{>{\raggedright}p{3.4cm} | r | r | r}
        \toprule
        Type & No. Episodes & No. steps & Hours\\
        \midrule
        Teleoperation & 2.8 K & 568 K & 15.8 \\
        \liftcloth{} & 13.3 K & 2.4 M & 66.0 \\
        \randwatch{} & 6.0 K & 1.2 M & 32.1 \\
        Total & 36.5 K & 6.9 M & 191.2 \\
        \bottomrule
        \end{tabular}
    \caption{\Deform{} dataset. \label{tab:deform8_dataset}}
\end{subtable}
\caption{Dataset statistics. Total includes off-task data not listed in individual rows, teleoperation and tasks \lift{}, \stack{}, \liftcloth{} partly overlap.}
\vspace{-5mm}
\label{tab:dataset}
\end{table}

To generate initial datasets for training we use a scripted policy called the \randwatch{}.
This policy moves the end effector to randomly chosen locations and opens and closes the gripper at random times.
When following this policy, the robot occasionally picks up or pushes the objects, but is typically just moving in free space.
This data not only serves to seed the initial iteration of learning, but removing it from the training datasets degrades performance of the final agents.

The datasets contain a significant number of teleoperated episodes. The majority are recorded via interactive  teleoperation (\autoref{sec:method-detail:teleop}), and thus require limited human intervention.
Only about \num{600} full teleoperated episodes correspond to the \lift{} or \stack{} tasks.

There are \num{894}, \num{1201}, and \num{585} sketched episodes for the \lift{},  \stack{} and \liftcloth{} tasks, respectively.
Approximately $90$\% of the episodes are used for training and $10$\% for validation.
The sketches are not obtained all at once, but accumulated over several iterations of the process illustrated in \autoref{fig:pull-figure}.
At the first iteration, the humans annotate randomly sampled demonstrations.
In next iterations, the annotations are usually done on agent data, and occasionally on demonstrations or random watcher data.
Note that only a small portion of data from \neverendingstorage{} is annotated.

\emph{Agent network architecture: }
The agent network is illustrated in \autoref{fig:agent_network}.
Each camera is encoded using a residual network followed by a spatial softmax keypoint encoder with \num{64} channels \cite{levine2016end}.
The spatial softmax layer produces a list of \num{64} $(x, y)$ coordinates.
We use one such list for each camera and concatenate the results.

Before applying the spatial softmax, we add noise from the distribution $\mathcal{U}[-0.1, 0.1]$ to the logits so that the network learns to concentrate its predictions, as illustrated with the circles in \autoref{fig:agent_network}.
Proprioceptive features are concatenated, embedded with a linear layer, layer-normalized~\cite{ba2016layer}, and finally mapped through a $\tanh$ activation. They are then appended to the camera encodings to form the joint input features.

The actor network $\pi(x)$ consumes these joint input features directly. The critic network $Q(x,a)$ additionally passes them through a linear layer, concatenates the result with actions passed through a linear layer, and maps the result through a linear layer with ReLU activations. The actor and critic networks each use
two layer-normalized LSTMs with $256$ hidden units.
Action outputs are further processed through a $\tanh$ layer placing them in the range $[-1, 1]$, and then re-scaled to their native ranges before being sent to the robot.

The agent for \lift{} and \stack{} tasks observes two cameras, a basket front left camera (\num{80x128}) and one of wrist-mounted wide angle cameras (\num{96x128}) (\autoref{fig:rig}).
The agent for \liftcloth{} uses an additional back left camera (\num{80x128}).

\emph{Reward network architecture: }
The reward network  is a non-recurrent residual network with a spatial softmax layer
\cite{levine2016end} as in the agent network architecture.
We also use the proprioceptive features as in the agent learning.
As the sketched values are in the range of $[0, 1]$, the reward network ends with a sigmoid non-linearity.

\begin{figure}[t]
	\centering
	\includegraphics[width=\columnwidth]{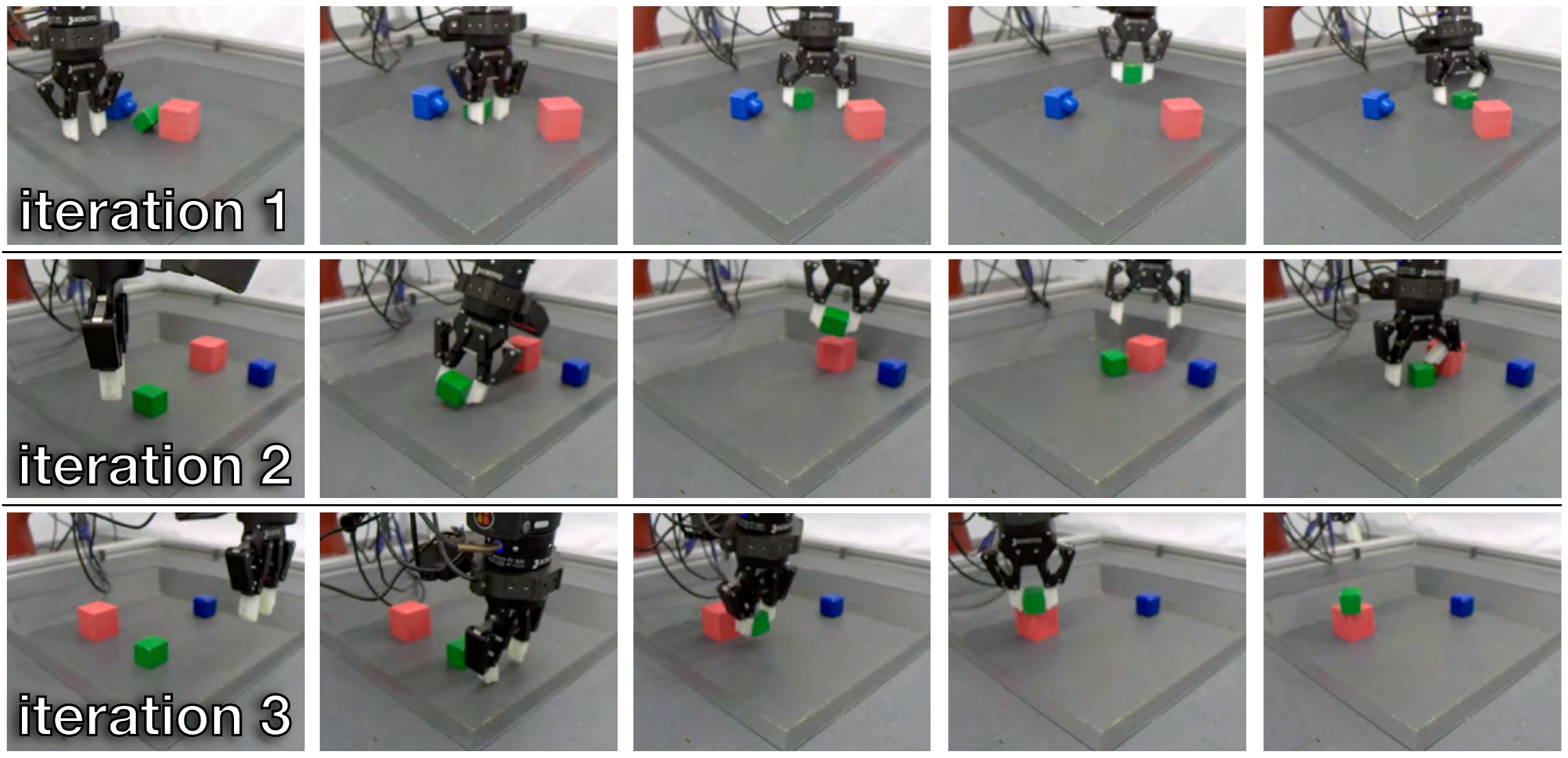}
	\caption{Iterative improvement of the agent on task \stack{}. Each iteration corresponds to a cycle through the steps as shown in \autoref{fig:pull-figure}. With more training data, the performance of agent improves. \label{fig:iterative-improvement}}
	\vspace{-5mm}
\end{figure}

\emph{Training: }
We train multiple RL agents in parallel and briefly evaluate the most promising ones on the robot.
Each agent is trained for \num{400}k update steps.
To further improve performance, we save all episodes from RL agents, and sketch more reward curves if necessary, and use them when training the next generation of agents.
We iterated this procedure \num{2}--\num{3} times
and at each iteration the agent becomes more successful and more robust.
Three typical episodes from three steps of improvement in \stack{} task are depicted in Fig.~\ref{fig:iterative-improvement}.
They correspond to agents trained using approximately $82$\%, $94$\% and $100$\% of the collected data.
In the first iteration, the agent could pick a green block, but drops it.
In the second iteration, the agent attempts stacking a green block on red, and only in the third iteration it succeeds in it.
Next, we report the performance of the final agents.

\emph{Evaluation: }
While the reward and policy are learned from data, we cannot assess their ultimate quality without running the agent on the real robot. That is, we need to evaluate whether our agents learned using the stored datasets transfer to the real robot.
As the agent is learned off-line, good performance on the real robot is a powerful indicator of generalization.

To this end, we conducted controlled evaluations on the physical robot with fixed initial conditions across different policies.
For the \lift{} and \stack{} datasets, we devise three different evaluation conditions with varying levels of difficulty:
\begin{enumerate}
    \item \emph{normal}: basic rectangular green blocks (well represented in the training data), large red objects close to the center;
    \item \emph{hard}: more diverse objects (less well represented in the training data), smaller red objects with diverse locations;
    \item \emph{unseen}: green objects that were never seen during training, large red objects.
\end{enumerate}

Each condition specifies \num{10} different initial positions of the objects (set by a human operator) as well as the initial pose of the robot (set automatically).  The \emph{hard} and \emph{unseen} conditions are especially challenging, since they require the agent to cope with novel objects and novel object configurations.

We use the same $3$ evaluation sets for both the \lift{} and \stack{} tasks.
To evaluate the \liftcloth{} task, we randomize the initial conditions at every trial.
As a quality metric, we measure the rate of successfully completed episodes, where success is indicated by a human operator.

\subsection{Results}

\begin{table}[t]
\begin{subtable}{\columnwidth}
    \centering
    \begin{tabular}{l|>{\raggedleft}p{0.7cm}>{\raggedleft}p{0.7cm}>{\raggedleft\arraybackslash}p{0.7cm}}
        \toprule
        Agent &  Normal &  Hard & Unseen \\
        \midrule
        \bf{Our approach} &  \bf{80\%} &  \bf{80\%} &  \bf{50\%} \\
        No random watcher data & \bf{80\%} &  70\% &  20\% \\
        Only lift data &  0\% &   0\% &   0\% \\
        Non-distributional RL &  30\% &  20\% &  10\% \\
        \bottomrule
    \end{tabular}
    \caption{\small\label{tab:policy_evaluation_lift_green} \lift{}}
\end{subtable}
\bigskip
\begin{subtable}{\columnwidth}
    \centering
    \begin{tabular}{l|>{\raggedleft}p{0.7cm}>{\raggedleft}p{0.7cm}>{\raggedleft\arraybackslash}p{0.7cm}}
        \toprule
        Agent &  Normal &  Hard & Unseen \\
        \midrule
        \bf{Our approach} &  \bf{60\%} &  \bf{40\%} &   \bf{40\%} \\
        No random watcher data &  50\% &  30\% &   30\% \\
        Only stacking data &  0\% &  10\% &   0\% \\
        Non-distributional RL & 20\% & 0\% & 0\% \\
        \bottomrule
    \end{tabular}
    \caption{\small\label{tab:policy_evaluation}\stack{}.}
\end{subtable}
\vspace{-2mm}
\caption{The success rate of our agent and ablations for a given  task in different difficulty settings. Recall that out agent is trained off-line.}
\vspace{-5mm}
\label{tab:policy_evaluation_all}
\end{table}

Results on the \rgb{} dataset are summarized in \autoref{tab:policy_evaluation_all}.
Our agent achieves a success rate of $80$\% for lifting and $60$\% for stacking.
Even with rarely seen objects positioned in adversarial ways, the agent is quite robust with success rates being $80$\% and $40$\%, respectively.
Remarkably, when dealing with objects that were never seen before, it can lift or stack them in $50$\% and $40$\% of cases (see \autoref{fig:film-strip} for examples of such behavior).
The success rate of our agent for the \liftcloth{} task in $50$ episodes with randomized initial conditions is $74$\%.

Our results compare favorably with those of Zhu~et~al.~\cite{zhu2018reinforcement}, where block lifting and stacking success rates are 64\% and 35\%. Note that these results are not perfectly comparable due to different physical setups, but we believe they provide some guidance.
Wulfmeier~et~al.~\cite{wulfmeier2019regularized} also attempted reward learning with the block stacking task. Instead of learning directly from pixels, they rely on QR-code state estimation for a fixed set of cubes, whereas our policies can handle objects of various shapes, sizes and material properties.
\citet{jeong2019self} achieve 62\% accuracy on block stacking (but with a fixed set of large blocks) using a sim2real approach with continuous $4$-DoF control. In contrast, we can achieve similar performance with a variety of objects and more complex continuous $6$-DoF control.

To understand the benefits of relabelling the past experience with learned reward functions, we conduct the ablations with fixed reward functions and varying training subsets for RL agents.
Firstly, we train the lifting (stacking) policy using only the lifting (stacking) episodes.
Using only task-specific data is interesting because the similarity between training data and target behavior is higher (\emph{i.e.}, the training data is more on-policy).
Secondly, we train an agent with access to data from all tasks, but no access to the \randwatch{} data.
As this data is unlikely to contain relevant to the task episodes, we want to know how much it contributes to the final performance.

\autoref{tab:policy_evaluation_all} show the results of these two ablations.
Remarkably, using only a task-specific dataset dramatically degrades the policy (its performance is $0$\% in almost all scenarios).
Random watcher data proves to be valuable as it contributes up to an additional $30$\% improvement, showing the biggest advantage in the hardest case with unseen objects.

We also evaluate the effect of distributional value functions.
Confirming previous findings in Atari \cite{agarwal2019striving}, the results in the last rows of \autoref{tab:policy_evaluation_all} show that distributional value functions are essential for good performance in batch RL.

\begin{figure}[t]
	\centering
	\includegraphics[width=\columnwidth]{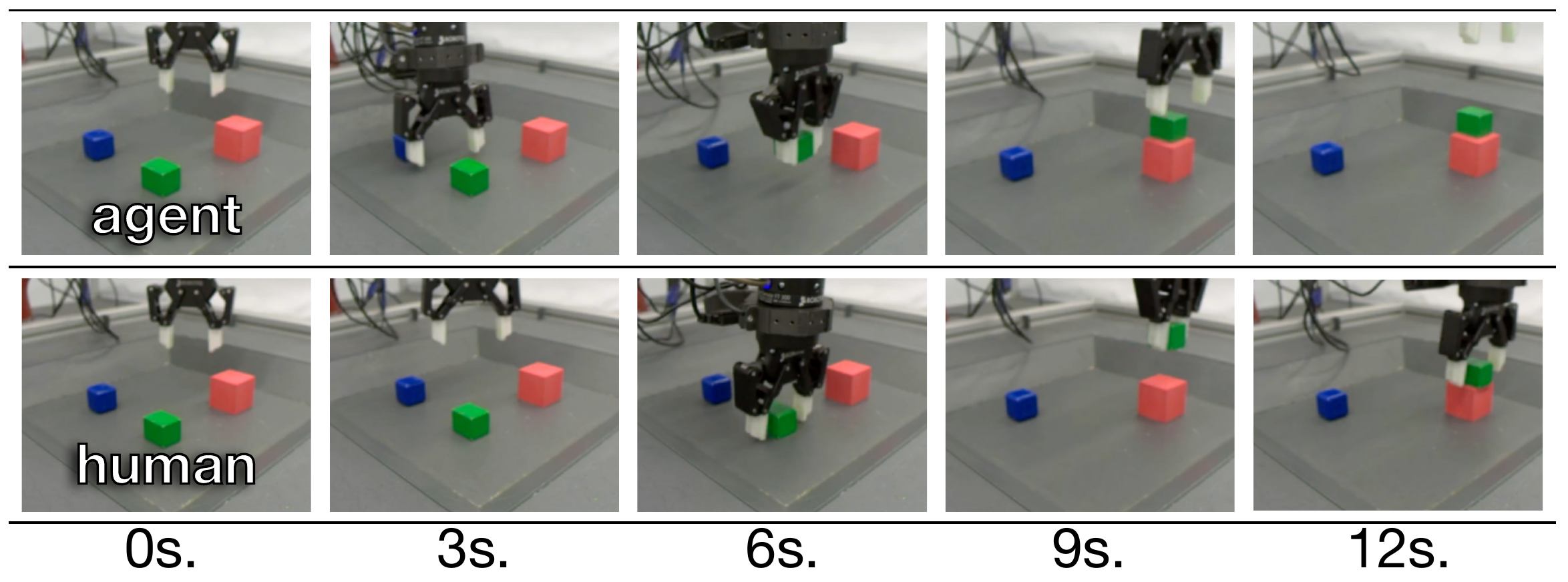}
	\caption{Agent vs human in \stack{} task. We show frames of an episode performed by an agent (top) and a human (bottom) after every $3$ seconds. The agent accomplishes the task faster than a human operator. \label{fig:agent_vs_human}}
	\vspace{-5mm}
\end{figure}

For qualitative results, we refer the reader to the accompanying video and \autoref{fig:film-strip} that demonstrate the robustness of our agents.
The robot successfully deals with adversarial perturbations by a human operator, stacking
several unseen and non-standard objects and lifting toys, such as a robot and a pony.
Our agents move faster and are more efficient compared to a human operator in some cases as illustrated in \autoref{fig:agent_vs_human}.

\subsection{Interactive Insertion}

\begin{figure}[t]
	\centering
	\includegraphics[width=\columnwidth]{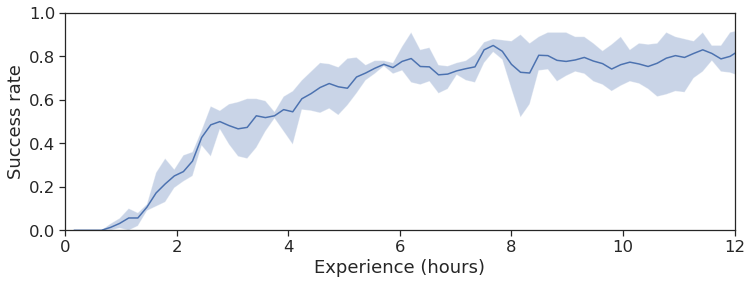}
	\caption{USB-insertion task success rate during the process of on-line training. It illustrates the rapid progress of training a robot to solve an industrially relevant task.}
	\vspace{-5mm}
	\label{fig:insertion_results}
\end{figure}

An alternative way to obtain a policy is to perform data collection, reward learning and policy learning in a tight loop.
Here, the human operator interactively refines the learned reward function on-line at the same time when a policy is learned.
In this experiment, the policy is learned from scratch without relying on historical data and batch RL, which is possible in less data-demanding applications.
In this section, we present an example of this approach applied to industrially relevant task: insert a USB key into a computer port.

We consider \num{6}-DoF velocity control.
The velocity actions are fed to a stateful safety controller, which uses a previously learned model to limit excess forces and a Mujoco inverse kinematics model to infer target joint velocities.
Episodes are set to last \num{15} seconds with \num{10} Hz control, for a total of \num{150} steps.
Both the policy and reward model use wrist camera images of size \num{84x84} pixels.

At the start of each episode, the robot position is set within a \num{6x6x6} cm region with \num{8.6}$^{\circ}$ rotation in each direction, and the allowed workspace is \num{8x8x15} cm with \num{17.2}$^{\circ}$ rotation. This is known to be significant amount of variation for such task.
Episodes are terminated with a discount of zero when the robot reached the boundary of the workspace.
For faster convergence, a smaller network architecture is chosen with \num{3} convolutional layers and \num{2} fully connected layers.
At the start of the experiment, \num{100} human demonstrations are collected and annotated with sketches.

This experiment is repeated $3$ times.
The average success rate of the agent as a function of time is shown in \autoref{fig:insertion_results}.
The agent reaches over \num{80}\% success rate within \num{8} hours.
During this time, the human annotator provides $65\pm10$ additional reward sketches.
This experiment demonstrates that it is possible to solve an industrial robot task from vision using human feedback within a single working day.

Two successful episodes of USB insertion are shown in Fig.~\ref{fig:film-strip} in two last rows.
In the first example the robot successfully inserts a key using only pixel inputs.
As only vision input is used during training and actions are defined with respect to the wrist frame, the resulting policy is robust to unseen positional changes.
In the second example, the agent (which is trained on the unperturbed state) can perform insertion despite moving the input socket significantly.

\section{Related Work}
\label{sec:relwork}

RL has a long history in robotics \cite{kober2013reinforcement, peters2008reinforcement, kalakrishnan2011learning, hafner2011reinforcement, levine2016end, levine2018learning, kalashnikov2018qtopt}.
However, applying RL in this domain inherits all the general difficulties of applying RL in the real world~\cite{dulac2019realworldRL}.
Most published works either rely on state estimation for a specific task, or work in a very limited regime to learn from raw observations.
These methods typically entail highly engineered reward functions.
In our work, we go beyond the usual scale of application of RL to robotics, learn from raw observations and without predefined rewards.

Batch RL trains policies from a fixed dataset and, thus, it is particularly useful in real-world applications like robotics.
It is currently an active area of research (see the work of \citet{lange2012batch} for an overview), with a number of recent works aimed at improving the stability~\cite{fujimoto2018off, jaques2019way, agarwal2019striving, kumar2019stabilizing}.

In the RL-robotics literature, QT-Opt~\cite{kalashnikov2018qtopt} is the closest approach to ours.
The authors collect a dataset of over \num{580000} grasps for several weeks with \num{7} robots.
They train a distributed Q-learning agent that shows remarkable generalization to different objects.
Yet, the whole system focuses on a single task: grasping.
This task is well-suited for reward engineering and scripted data collection policies.
However, these techniques are not easy to design for many tasks and, thus, relying on them limits the applicability of the method.
In contrast, we collect the diverse data and we learn the reward functions.

Learning reward functions using inverse RL~\cite{ng2000algorithms} achieved tremendous success \cite{finn2016guided, ho2016generative, li2017infogail,fu2017learning, merel2017learning,zhu2018reinforcement,baram2017end}.
This class of methods works best when applied to states or well-engineered features.
Making it work for high-dimensional input spaces, particularly raw pixels, remains a great challenge.

Learning from preferences has a long history~\cite{Thurstone1927a,Mosteller1951remarks,Feinberg1976loglinear,Stern1990a,Chu2005preference,Joachims2007evaluating}.
Interactive learning and optimization with human preferences dates back to works at the interface of machine learning and graphics \cite{brochu2007active,brochu2010interactive}.
Preference elicitation is also used for reward learning in RL~\cite{Strens2003policy,wirth2017a}.
It can be done by whole episode comparisons~\cite{Akrour2012april,Schoenauer2014programming, brown2019extrapolating, dorsa2017active} or shorter clip comparisons~\cite{christiano2017deep,ibarz2018reward}.
A core challenge is to engineer methods that acquire many preferences with as little user input as possible~\cite{koyama2017sequential}.
To deal with this challenge, our reward sketching interface allows perceptual reward learning~\cite{sermanet2017unsupervised} from any, even unsuccessful trajectories.

Many works in robotics choose to learn from demonstrations to avoid hard exploration problems of RL.
For example, supervised learning to mimic demontrations is done in BC~\cite{pomerleau1989alvinn, rahmatizadeh2018vision}.
However, BC requires high-quality consistent demonstrations of the target task and as such, it cannot benefit from heterogeneous data.
Moreover, BC policies generally cannot outperform the human demonstrator.
Demonstrations could be also used in RL~\cite{nair2018overcoming, rajeswaran2017learning} to address the exploration problem.
As in prior works~\cite{vecerik2017leveraging, pohlen2018observe, vecerik2019practical}, we use demonstrations as part of the agent experience and train with temporal difference learning in a model-free setting.

Several recent large-scale robotic datasets were released recently to advance the data-driven robotics.
Roboturk~\cite{mandlekar2018roboturk} collects crowd-sourced demonstrations for three tasks with the mobile platform. The dataset is used in the experiments with online RL.
MIME~\cite{pratyusha18mime} dataset contains both human and robot demonstrations for \num{20} diverse tasks and its potential is tested in the experiments with BC and task recognition.
RoboNet~\cite{dasari2019robonet} database focuses on transferring the experience across objects and robotic platforms. The large-scale collection of the data is possible thanks to scripted policies. The strength of this dataset is evaluated in action-conditioned video prediction and in action prediction.
Our dataset~\cite{outdataset} is collected with demonstrations, scripted policies as well as learned policies.
This paper is the first to show how to efficiently label such datasets with rewards and how to apply batch RL to such challenging domains.

\section{Conclusions}
\label{sec:conclusions}

We have proposed a new data-driven approach to robotics.
Its key components include a method for reward learning, retrospective reward labelling and batch RL with distributional value functions.
A significant amount of engineering and innovation was required to implement this at the present scale.
To further advance data-driven robotics, reward learning and batch RL, we release the large datasets~\cite{outdataset} from NeverEnding Storage and canonical agents~\cite{hoffman2020acme}.

We found that reward sketching is an effective way to elicit reward functions, since humans are good at judging progress toward a goal.
In addition, the paper also showed that storing robot experience over a long period of time and across different tasks allows to efficiently learn policies in a completely off-line manner.
Interestingly, diversity of training data seems to be an essential factor in the success of standard state-of-the-art RL algorithms, which were previously reported to fail when trained only on expert data or the history of a single agent~\cite{fujimoto2018off}.
Our results across a wide set of tasks illustrate the versatility of our data-driven approach. In particular, the learned agents showed a significant degree of generalization and robustness.

This approach has its limitations.
For example, it involves a human-in-the-loop during training which implies additional cost.
The reward sketching procedure is not universal and other strategies might be needed for different tasks.
Besides, the learned agents remain sensitive to significant perturbations in the setup.
These open questions are directions for future work.

\section*{Acknowledgments}
We would like to thank all the colleagues at DeepMind who teleoperated the robot for data collection.

\clearpage
{\small
\bibliographystyle{plainnat}
\bibliography{refs}
}

\end{document}